\documentclass[12pt,letterpaper]{article}
\usepackage{float}
\usepackage[a4paper, total={7in, 10in}]{geometry}

\usepackage{graphicx}
\usepackage{helvet}
\usepackage{authblk}
\usepackage{hyperref}
\usepackage{amsmath} 
\usepackage{amssymb} 
\usepackage{orcidlink} 
\usepackage{float} 
\restylefloat{figure}
\usepackage[super,comma,sort&compress]{natbib}
\usepackage[right]{lineno} \linenumbers
\usepackage[title]{appendix}%
\usepackage[utf8]{inputenc} % allow utf-8 input
\usepackage[T1]{fontenc}    % use 8-bit T1 fonts
\usepackage{url}            % simple URL typesetting
\usepackage{booktabs}       % professional-quality tables
\usepackage{amsfonts}       % blackboard math symbols
\usepackage{nicefrac}       % compact symbols for 1/2, etc.
\usepackage{microtype}      % microtypography
\usepackage{xcolor}         % colors
\usepackage{multirow}
\usepackage{threeparttable}
\usepackage{amsmath}
\usepackage{subcaption}
\usepackage{makecell}
\usepackage{placeins}
\usepackage[utf8]{inputenc} % 支持 UTF-8 编码
\usepackage{booktabs}      % 用于生成专业的三线表     % 用于按小数点对齐数字
\usepackage{caption}       % 用于表格标题
\usepackage{tabularx}
\usepackage{listings}    % 引入 listings 宏包
\usepackage{pifont}
\usepackage{array}
\usepackage{longtable}
\usepackage{booktabs} % 保留单元格换行功能

\theadset{\bfseries}

\makeatletter
\renewcommand{\maketitle}{\bgroup\setlength{\parindent}{0pt}
\begin{flushleft}
  \textbf{\@title}
  
  \@author
\end{flushleft}\egroup}
\makeatother

\title{Reconstructing 12-Lead ECG from 3-Lead ECG using Variational Autoencoder to Improve Cardiac Disease Detection of Wearable ECG Devices}
\date{}

\author[1,2]{Xinyan Guan}
\author[3,4]{Yongfan Lai}
\author[1,2,3,4]{Jiarui Jin}
\author[1,2]{Jun Li}
\author[1]{Haoyu Wang}
\author[5]{Qinghao Zhao}
\author[6]{Deyun Zhang}
\author[6]{Shijia Geng}
\author[1,2,*,\orcidlink{0000-0001-7521-5127}]{Shenda Hong}

\affil[1]{National Institute of Health Data Science, Peking University, Beijing, China}
\affil[2]{Institute for Artificial Intelligence, Peking University, Beijing, China}
\affil[3]{State Key Laboratory of General Artificial Intelligence, Beijing 100871, China}
\affil[4]{School of Intelligence Science and Technology, Peking University, Beijing 100871, China}
\affil[5]{Department of Cardiology, Peking University People’s Hospital, Beijing 100044, China}
\affil[6]{HeartVoice Medical Technology, Hefei 230088, China}

\affil[*]{Correspondence: hongshenda@pku.edu.cn}

\begin{document}

\maketitle

\section*{ABSTRACT}
Twelve-lead electrocardiograms (ECGs) are the clinical gold standard for cardiac diagnosis, offering comprehensive spatial coverage of the heart necessary for detecting conditions such as myocardial infarction (MI). However, their lack of portability limits continuous and large-scale deployment. In contrast, three-lead ECG systems are widely used in wearable devices due to their simplicity and mobility, but they often fail to capture pathologies localized in unmeasured regions. To bridge this gap, we propose WearECG, a Variational Autoencoder (VAE) method that reconstructs 12-lead ECGs from three leads (II, V1, V5). Our model includes architectural improvements to better capture temporal and spatial dependencies in ECG signals. We evaluate generation quality using MSE, MAE, and Fréchet Inception Distance (FID), and assess clinical validity via a Turing test with expert cardiologists. To further validate the diagnostic utility, we fine-tune ECGFounder—a large-scale pretrained ECG model—on a multi-label classification task involving over 40 cardiac conditions, including 6 different myocardial infarction locations using both real and generated signals. Experiments on the MIMIC dataset show that our method produces physiologically realistic and diagnostically informative signals, with robust performance in downstream tasks. This work demonstrates the potential of generative modeling for ECG reconstruction and its implications for scalable, low-cost cardiac screening.

\section*{KEYWORDS}
AI in Cardiology, ECG, Wearable Devices, Variational Autoencoder
%%%  Include up to 10 keywords, separated by commas. 
%%%  Keywords entered in EM are not carried over; only 
%%%  keywords included in the main text will be used 
%%%  in the final article metadata. Please note that 
%%%  for some journals, keywords are chosen by the editors.

\section*{INTRODUCTION}
As the leading global cause of mortality, cardiovascular disease (CVD) ~\cite{Nabel2003,Roth2018}necessitates reliable diagnostic tools, among which the 12-lead electrocardiogram (ECG)—performed over 300 million times annually—has become fundamental. The 12-lead ECG is considered the gold standard for non-invasive cardiac assessment owing to its ability to provide comprehensive electrical activity mapping of the heart. However, despite its diagnostic value, the standard 12-lead ECG lacks portability. It is not suitable for continuous, ambulatory monitoring. As a result, many cardiovascular events occur without timely detection or intervention, contributing to the high mortality associated with conditions such as myocardial infarction and arrhythmia. To address this gap, wearable ECG devices have emerged as a promising solution, enabling long-term cardiac monitoring in daily life and facilitating early detection of critical events. Accordingly, the development of user-centric devices for pervasive ECG signal acquisition has emerged as a central goal in both academic research and commercial innovation, encompassing patch-type systems ~\cite{Turakhia2013,Lai2020,Liu2021}, smartwatches ~\cite{Tison2018,Bumgarner2018,Perez2019}, and armband-based solutions ~\cite{Rachim2016,Li2021,Lazaro2020}.

Given that portable devices typically capture only a few leads, researchers have spent the past thirty years exploring how to reconstruct the full 12-lead ECG by exploiting the inherent inter-lead correlations. While initial methods were predominantly based on linear transformations, the advent of artificial intelligence has paved the way for more advanced and effective reconstruction strategies.

Several prior studies have explored the reconstruction of standard 12-lead ECG signals using conventional techniques. Early approaches leveraged linear transformation matrices to capture inter-lead correlations~\cite{nelwan2000minimal, maheshwari2016frank}, while others employed temporal modeling strategies~\cite{lee2016state, zhu2018lightweight}. However, many of these methods depend on patient-specific algorithms~\cite{tsouri2014patient}, which significantly limit their generalizability. Given that the relationships among ECG leads vary across individuals, such hand-crafted approaches often fail to capture the nonlinear and dynamic nature of inter-lead dependencies. These limitations have motivated researchers to explore data-driven alternatives capable of learning more flexible and generalizable mappings from incomplete inputs to full 12-lead signals.

Supervised deep learning models have gained increasing attention in this context, particularly due to their ability to leverage large-scale annotated ECG datasets. Nejedly et al.\cite{nejedly2021} introduced an ensemble learning framework that integrates residual convolutional neural networks (CNNs) with an attention mechanism, achieving strong classification performance. Building on this, Gundlapalle and Acharyya\cite{gundlapalle2022} combined CNNs and LSTM modules to model both spatial and temporal patterns in single-lead to 12-lead reconstruction. Garg et al.\cite{garg2023} proposed a modified Attention U-Net for reconstructing multiple leads from a single input, while Chen et al.\cite{chen2024} introduced a Multi-Channel Masked Autoencoder capable of generalizing across diverse input configurations. These approaches demonstrate the potential of discriminative models in ECG reconstruction, but often require abundant labeled data and still face limitations when input information is highly sparse.

To address these challenges, recent work has turned to generative models that can synthesize plausible ECG signals even under severely under-constrained input conditions. Lee et al.\cite{lee2020synthesis} proposed a conditional generative adversarial network (CGAN) aligned with R-peaks to reconstruct precordial leads (V1–V6) from limb-lead inputs, achieving high fidelity in terms of SSIM ($\approx$ 0.92) and PRD ($\approx$ 7.21\%). Seo et al.\cite{seo2022multiple} introduced a WGAN-based architecture featuring a U-Net generator and CNN discriminator to generate multiple leads from a single-lead input, evaluated using Fréchet Distance and MSE. More recently, Joo et al.~\cite{joo2023twelve} developed EKGAN, a dual-generator GAN framework with a 1D U-Net discriminator, which demonstrated superior performance in reconstructing full 12-lead ECGs from Lead I, as validated by both quantitative metrics and expert cardiologists. These generative approaches offer a promising direction for robust ECG reconstruction in low-resource or real-world deployment scenarios.

Single-lead ECG inputs are widely used in portable and home-monitoring devices but provide limited spatial information, which restricts their clinical interpretability, particularly for detecting regional abnormalities such as myocardial infarctions. While three-lead inputs capture orthogonal cardiac directions and enable more robust reconstruction, existing models like Mason et al.’s stacked ResNet~\cite{mason2024ai} suffer from architectural complexity that may limit clinical applicability. Moreover, most current generative methods rely on GAN frameworks, which are notorious for training instability, mode collapse, and sensitivity to hyperparameters, making them challenging to optimize and deploy in real-world clinical settings. 

To address these limitations, we explore generating 12-lead ECGs from three-lead inputs using a carefully designed VAE framework, incorporating multi-scale residual blocks, attention-enhanced bottlenecks, and a structured latent distribution. The encoder-decoder architecture leverages downsampling with residual convolutional blocks and group normalization, while KL-regularized latent sampling ensures stable training. The learned latent space supports physiologically plausible reconstructions, even under severely limited input leads.Therefore, the contributions in this study are as follows:
\begin{itemize}
\item We develop a VAE-based model for reconstructing 12-lead ECGs from three commonly used leads: II, V1, and V5.
\item Our model achieves high-fidelity reconstruction, with an overall MSE of 0.00100, MAE of 0.01782, and FID of 12.64 on the test set.
\item Beyond signal-level evaluation, we assess the clinical validity of the reconstructed ECGs on multiple diagnostic classification tasks (e.g., acute MI/STEMI, atrial fibrillation, atrial flutter, left axis deviation), achieving consistently high AUROC scores.

\end{itemize}

\section*{RESULTS}
To comprehensively evaluate the performance of our proposed model, we conducted a multi-level assessment spanning signal-level reconstruction quality, feature-level realism, and diagnostic utility. At the signal level, we quantitatively measured the similarity between generated and reference ECGs using metrics such as Mean Squared Error (MSE), Mean Absolute Error (MAE), and Fréchet Inception Distance (FID), capturing both numerical accuracy and perceptual closeness. To evaluate feature-level fidelity, we performed a blinded Turing test with three cardiology experts, who was tasked with distinguishing real from synthetic ECG signals, providing insight into the clinical plausibility of our generated data. Finally, at the diagnostic level, we fine-tuned a classification head on top of a frozen ECGFounder backbone using the generated ECGs and evaluated performance across multiple cardiac conditions by AUROC, reflecting how well the synthetic signals preserve disease-specific characteristics. This comprehensive evaluation framework ensures a holistic understanding of the model’s ability to generate ECGs that are accurate, realistic, and diagnostically meaningful.

\subsection*{Method Overview}

% \subsection*{Datasets}
We utilized the MIMIC-IV-ECG matched subset to train and validate our ECG generation model. This subset comprises approximately 800,000 ten-second 12-lead diagnostic ECG recordings from approximately 160,000 unique patients, collected at Beth Israel Deaconess Medical Center between 2008 and 2019. Each ECG is sampled at 500 Hz, stored in standard WFDB format, and is linked to patient metadata (demographics, RR interval) via shared \texttt{subject\_id} and \texttt{study\_id} with the MIMIC-IV Clinical Database \cite{mimiciv_ecg}.

When available, machine-generated summary measurements (e.g., RR intervals, QRS onset/end) and de-identified cardiologist text reports are provided for the same recordings. Identifiers and timestamps are privacy-protected using HIPAA-compliant de-identification, with shifted but internally consistent date-times enabling time-alignment across ECGs and other clinical events \cite{mimiciv_ecg}.

Because nearly all diagnostic ECGs for MIMIC-IV clinical patients are included, this dataset allows us to link ECG waveforms to hospitalization data such as admission, discharge diagnoses, age, sex, and clinical outcomes \cite{mimiciv_ecg}. We retrieved age and sex information via the Clinical Database and calculated heart rate from RR interval records in the ECG measurements.

\begin{figure}[H]
  \centering
  \includegraphics[width=0.8\linewidth]{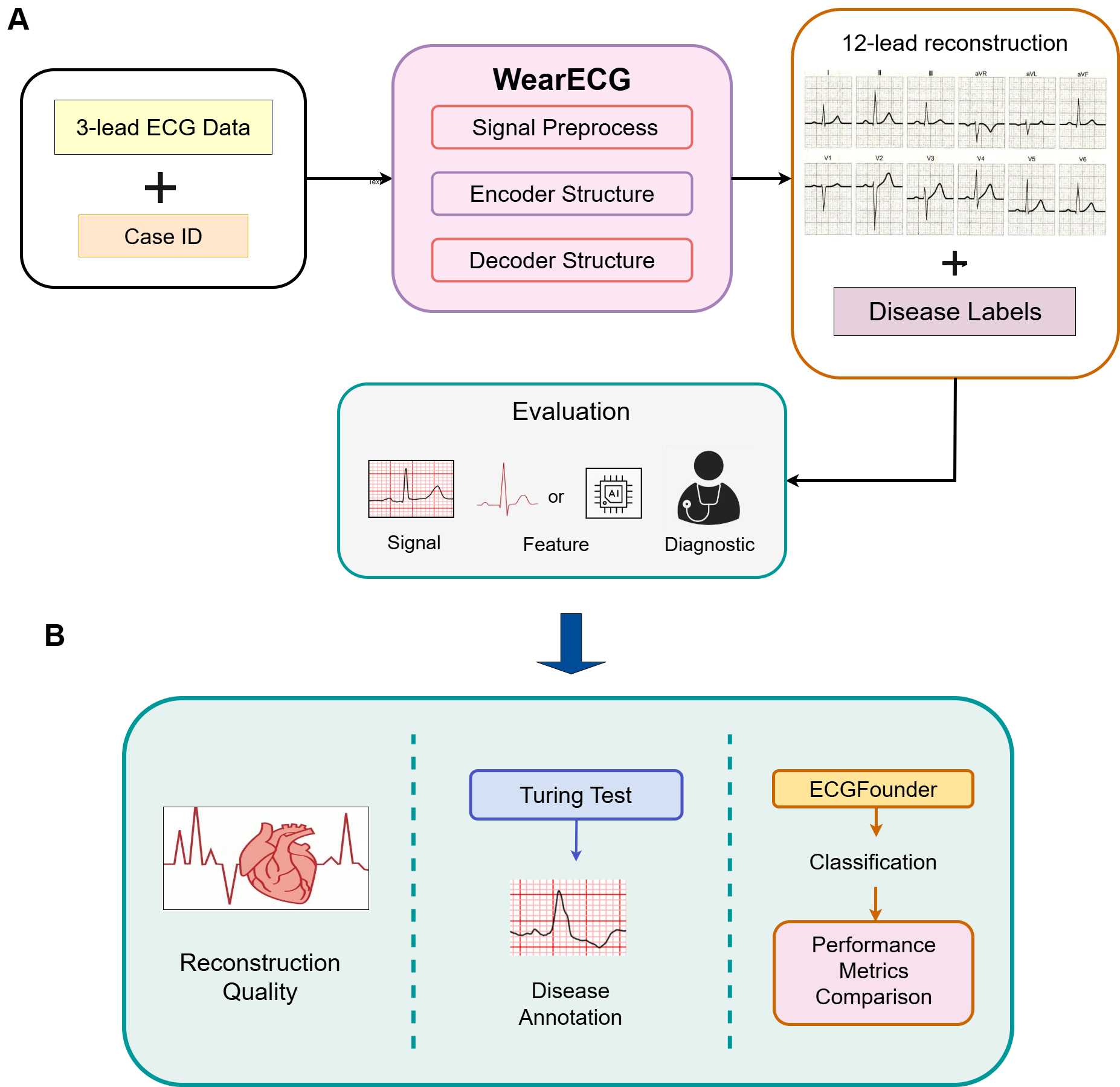}
  \caption{
    Overview of our framework. 
    \textbf{(A)} Three-lead ECG signals are fed into a generative model to reconstruct full 12-lead ECGs.
    \textbf{(B)} The reconstructed ECGs are evaluated via multiple strategies, including signal-level metrics, 
    fine-tuned disease classification using ECGFounder and cardiologist-involved Turing test.
  }
  \label{fig:overview}
\end{figure}

In the reconstruction task, we selected leads II, V1, and V5 because their spatial positions are mutually perpendicular, forming an orthogonal triad that effectively captures the multidimensional electrical activity of the heart from different anatomical planes. Lead II primarily reflects the inferior aspect, V1 focuses on the right ventricular and septal region, and V5 captures the lateral wall of the left ventricle. This complementary spatial coverage allows for a more accurate and comprehensive reconstruction of the full 12-lead ECG.

We applied an improved Variational Autoencoder (VAE) architecture tailored for ECG signal characteristics to reconstruct the missing leads. The enhanced model architecture facilitates better feature extraction and signal generation quality, preserving key diagnostic patterns. The reconstructed 12-lead ECGs were then used as input for the ECGFounder model in downstream disease classification tasks, achieving high AUROC scores and demonstrating that the synthetic signals retain clinically relevant information.

The diagram below illustrates the spatial relationship among these three leads, visually highlighting why this specific lead selection provides rich, complementary cardiac electrical information essential for effective reconstruction.
\begin{figure}[H]
    \centering
    \includegraphics[width=0.7\linewidth]{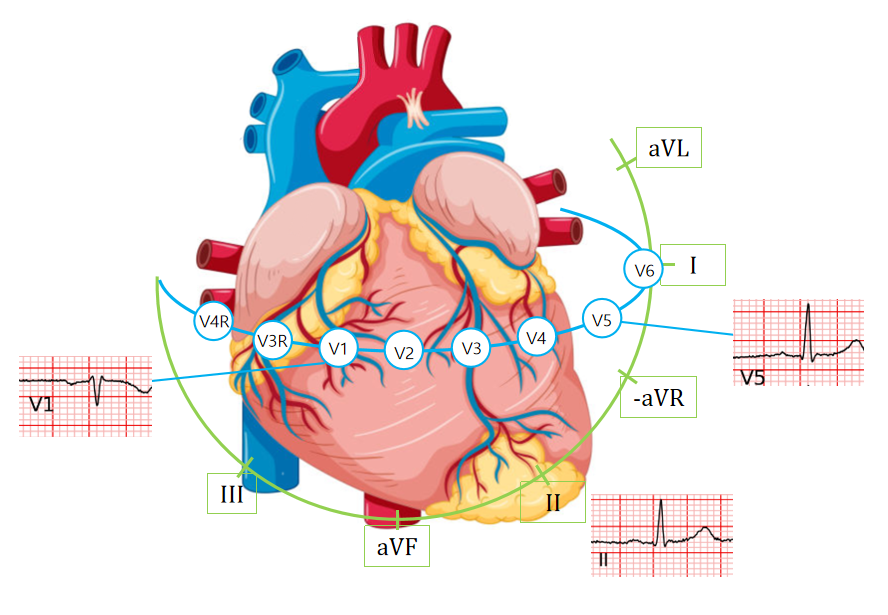}
    \caption{Spatial relationship among leads II V1 V5}
    \label{fig:leads}
\end{figure}
\subsection*{WearECG can Generate High Fidelity ECG}
\begin{table}[H]
\centering
\caption{Signal-level evaluation metrics for all experiments on the MIMIC dataset with different input leads.}
\label{tab:combined_metrics}
\small
\begin{tabular}{ccccc}
\toprule
Lead Name & Lead Index & MSE & MAE & FID Score (if available) \\
\midrule
\multicolumn{5}{c}{\textbf{Main experiment (Input leads: II, V1, V5)}} \\
\midrule
I    & 0  & 0.00075 & 0.01691 &  \\
III  & 2  & 0.00087 & 0.01833 &  \\
aVR  & 3  & 0.00061 & 0.01425 &  \\
aVL  & 4  & 0.00067 & 0.01570 &  \\
aVF  & 5  & 0.00073 & 0.01568 &  \\
V2   & 7  & 0.00126 & 0.02051 &  \\
V3   & 8  & 0.00135 & 0.01984 &  \\
V4   & 9  & 0.00138 & 0.01997 &  \\
V6   & 11 & 0.00140 & 0.01916 &  \\
\midrule
\multicolumn{2}{c}{Overall Metrics} & 0.00100 & 0.01783 & 11.34 \\
\midrule
\multicolumn{5}{c}{\textbf{Comparative experiment (Input leads: I, II, V3)}} \\
\midrule
I    & 0  & 0.00085 & 0.01826 &  \\
III  & 2  & 0.00061 & 0.01468 &  \\
aVR  & 3  & 0.00067 & 0.01571 &  \\
aVL  & 4  & 0.00070 & 0.01578 &  \\
aVF  & 5  & 0.00162 & 0.02073 &  \\
V2   & 7  & 0.00125 & 0.02047 &  \\
V3   & 8  & 0.00148 & 0.02071 &  \\
V4   & 9  & 0.00150 & 0.02088 &  \\
V6   & 11 & 0.00149 & 0.01981 &  \\
\midrule
\multicolumn{2}{c}{Overall Metrics} & 0.00113 & 0.01856 & 11.58 \\
\midrule
\multicolumn{5}{c}{\textbf{Comparative experiment (Input lead: I)}} \\
\midrule
II   & 1  & 0.00076 & 0.01826 &  \\
III  & 2  & 0.00081 & 0.01565 &  \\
aVR  & 3  & 0.00067 & 0.01572 &  \\
aVL  & 4  & 0.00083 & 0.01575 &  \\
aVF  & 5  & 0.00182 & 0.02174 &  \\
V1   & 6  & 0.00147 & 0.02348 &  \\
V2   & 7  & 0.00143 & 0.02048 &  \\
V3   & 8  & 0.00158 & 0.02171 &  \\
V4   & 9  & 0.00172 & 0.02289 &  \\
V5   & 10 & 0.00159 & 0.01880 &  \\
V6   & 11 & 0.00172 & 0.01980 &  \\
\midrule
\multicolumn{2}{c}{Overall Metrics} & 0.00131 & 0.01939 & 12.52 \\
\bottomrule
\end{tabular}
\end{table}

To quantitatively evaluate the fidelity of generated ECG signals, we adopt three objective signal-level metrics: Mean Absolute Error (MAE), Mean Squared Error (MSE), and Fréchet Inception Distance (FID). These metrics assess both point-wise and distributional similarity between the generated signals and real ECGs.

Let $\mathbf{r}_{\text{ECG}} \in \mathbb{R}^{T \times C}$ and $\mathbf{g}_{\text{ECG}} \in \mathbb{R}^{T \times C}$ denote the real and generated ECG signals respectively, where $T$ is the number of time points and $C$ is the number of channels. The MAE and MSE are defined as:

\begin{equation}
\text{MAE}(\mathbf{r}_{\text{ECG}}, \mathbf{g}_{\text{ECG}}) = \frac{1}{T \cdot C} \sum_{t=1}^{T} \sum_{c=1}^{C} \left| \mathbf{r}_{t,c} - \mathbf{g}_{t,c} \right|
\end{equation}

\begin{equation}
\text{MSE}(\mathbf{r}_{\text{ECG}}, \mathbf{g}_{\text{ECG}}) = \frac{1}{T \cdot C} \sum_{t=1}^{T} \sum_{c=1}^{C} \left( \mathbf{r}_{t,c} - \mathbf{g}_{t,c} \right)^2
\end{equation}

MAE captures the average magnitude of signal reconstruction error, while MSE penalizes larger deviations more heavily due to squaring. Lower MAE and MSE values indicate more accurate signal-level reconstructions.

To further evaluate the distributional similarity in the feature space, we adopt the Fréchet Inception Distance (FID) metric, which measures the distance between the feature distributions of generated and real ECGs. The features are extracted from a pretrained encoder (e.g., ECGFounder), and the FID is calculated as:

\begin{equation}
\text{FID} = \| \mu_r - \mu_g \|_2^2 + \text{Tr}\left( \Sigma_r + \Sigma_g - 2 (\Sigma_r \Sigma_g)^{1/2} \right)
\end{equation}

Here, $\mu_r$, $\Sigma_r$ and $\mu_g$, $\Sigma_g$ denote the mean and covariance matrices of feature embeddings for real and generated samples, respectively. A lower FID indicates that the synthetic signals are more similar to real ECGs in terms of high-level representations.
 
For each experimental setup, Tables~\ref{tab:combined_metrics},  list the MAE and MSE values per lead alongside the overall averaged scores, providing a comprehensive view of reconstruction quality across all leads.

In the main experiment using leads II, V1, and V5 as input, the per-lead MAE and MSE values are consistently low, with the MAE ranging from approximately 0.00061 to 0.00140. The overall MAE and MSE are 0.00100 and 0.01782, respectively. The corresponding FID score of 12.64 further suggests that the reconstructed signals closely resemble the real ECG distribution in the feature space. Clinically, this lead combination offers broad spatial coverage across the frontal and horizontal planes, effectively capturing both limb and precordial information—making it a common configuration in arrhythmia and myocardial infarction screening.

To validate the robustness of our method, we compared it with two alternative configurations: leads I, II, and V3, and the single-lead I setup. Both setups also yield strong performance, with overall MAE/MSE scores of 0.00113/0.01856 and 0.00112/0.01863, respectively. These results highlight that while our generative framework is robust to variations in input leads, carefully selecting leads with high diagnostic value further enhances reconstruction accuracy and practical applicability.

\subsection*{WearECG can Enhance the Performance of AI-ECG Diagnosis}
To rigorously evaluate the diagnostic capacity of the generated 12-lead ECG signals, we performed a downstream multi-label classification task using ECGFounder\cite{li2024ecgfounder}, a leading foundation model pretrained on over 10 million clinical ECG records. This task directly probes whether the generated signals retain sufficient pathological information to support real-world clinical decision-making. Specifically, the model was tasked with predicting approximately 40 cardiovascular conditions—including arrhythmias, conduction abnormalities, myocardial infarction, and chamber enlargements—using multi-hot encoded labels that reflect the frequent co-occurrence of cardiac diseases. This setting imposes a high bar for diagnostic fidelity, as successful classification requires the presence of nuanced, disease-specific signal features.

To isolate the diagnostic value of the generated signals, we froze the pretrained ECGFounder encoder and trained a lightweight classification head solely on synthetic ECGs. Prior to training, all signals were normalized on a per-lead basis via Z-score standardization. Performance was comprehensively evaluated using AUROC, sensitivity, specificity, and optimal threshold per condition, providing insights into both global and condition-specific diagnostic utility. Strong performance in this task indicates that our generation framework does more than mimic surface waveform morphology—it embeds clinically meaningful patterns that generalize to real-world diagnostic objectives, underscoring the potential value of synthetic ECGs in training, benchmarking, and augmenting downstream clinical models.

\begin{longtable}{p{4.5cm}cccc}
\caption{AUROC comparison across different ECG lead configurations on MIMIC Dataset: WearECG (ours), original 12-lead, 3-lead only, and 1-lead only.} \\
\label{tab:synthetic_eval} \\
\toprule
Disease & WearECG (ours) & original 12-lead & 3-lead only & 1-lead only \\
\midrule
\endfirsthead

\multicolumn{5}{c}{{\bfseries Table \thetable\ continued}} \\
\toprule
Disease & WearECG (ours) & original 12-lead & 3-lead only & 1-lead only \\
\midrule
\endhead

\bottomrule
\endfoot

NORMAL & 0.7958 & 0.8468 & 0.7753 & 0.8071 \\
SINUS RHYTHM & 0.9247 & 0.9393 & 0.9262 & 0.9397 \\
SINUS BRADY & 0.9882 & 0.9881 & 0.9878 & 0.9848 \\
AFIB & 0.9751 & 0.9773 & 0.9718 & 0.9700 \\
SINUS TACHY & 0.9910 & 0.9928 & 0.9928 & 0.9914 \\
LAXIS DEV & 0.8609 & 0.8562 & 0.9269 & 0.2827 \\
PVC & 0.9551 & 0.9391 & 0.9389 & 0.8539 \\
RBBB & 0.9411 & 0.9588 & 0.9367 & 0.8632 \\
LAE & 0.7290 & 0.8103 & 0.6391 & 0.6578 \\
PAC & 0.9722 & 0.9570 & 0.9530 & 0.9573 \\
PSVC & 0.8785 & 0.8852 & 0.8707 & 0.8614 \\
LBBB & 0.7519 & 0.7448 & 0.7873 & 0.7224 \\
LVH & 0.6881 & 0.7666 & 0.7129 & 0.5812 \\
QT SHORT & 0.7406 & 0.7741 & 0.6956 & 0.5987 \\
QT LONG & 0.8025 & 0.8224 & 0.8076 & 0.7977 \\
AFLUTTER & 0.9088 & 0.9068 & 0.8962 & 0.8785 \\
SINUS ARR & 0.9418 & 0.9533 & 0.9278 & 0.8961 \\
LAFB & 0.9065 & 0.9031 & 0.9603 & 0.5420 \\
RAXIS DEV & 0.8588 & 0.8540 & 0.7259 & 0.5496 \\
ECTOPIC ATR & 0.7621 & 0.8204 & 0.7989 & 0.8932 \\
SHORT PR & 0.7600 & 0.8216 & 0.7818 & 0.7248 \\
REPOL ABN & 0.6019 & 0.6534 & 0.5176 & 0.5189 \\
RAE & 0.6219 & 0.7086 & 0.7660 & 0.8165 \\
VOLT CRIT LVH & 0.6310 & 0.6159 & 0.7123 & 0.5627 \\
LPFB & 0.7822 & 0.8294 & 0.7759 & 0.3841 \\
1ST AVB & 0.8752 & 0.9282 & 0.7953 & 0.8537 \\
RVH & 0.8047 & 0.7927 & 0.7599 & 0.6064 \\
ACUTE MI/STEMI & 0.8427 & 0.8734 & 0.6794 & 0.6870 \\
SVT & 0.8374 & 0.8540 & 0.8130 & 0.8265 \\
VT & 0.8815 & 0.8925 & 0.8854 & 0.8740 \\
EARLY REP & 0.7927 & 0.7954 & 0.6801 & 0.7243 \\
WPW & 0.8831 & 0.9050 & 0.8306 & 0.8558 \\
ACUTE & 0.8469 & 0.8774 & 0.7280 & 0.6922 \\
ACUTE MI & 0.8437 & 0.8720 & 0.6971 & 0.7213 \\
SVC & 0.8785 & 0.8853 & 0.8718 & 0.8620 \\
2:1 AV COND & 0.7428 & 0.6904 & 0.6936 & 0.6256 \\

\textbf{Macro-AUC} & \textbf{0.8333} & \textbf{0.8465} & \textbf{0.7837} & \textbf{0.7545} \\

\end{longtable}
 
\begin{longtable}{p{4.5cm}cccc}
\caption[Regional MI AUROC]{Regional MI classification on MIMIC Dataset (AUROC)} \\[-1.5ex] % -1.5ex 可调整高度
\label{tab:MI_classification} \\
\toprule
Region & WearECG & 1-lead only & 3-lead only & Original 12-lead \\
\endfirsthead

\hline
Region & WearECG & 1-lead only & 3-lead only & Original 12-lead \\
\hline
\endhead

Anterior       & 0.8683 & 0.7966 & 0.8397 & 0.8728 \\
Anterolateral  & 0.8880 & 0.7979 & 0.8297 & 0.8894 \\
Anteroseptal   & 0.9378 & 0.7676 & 0.8624 & 0.9410 \\
Inferior       & 0.8067 & 0.8180 & 0.8055 & 0.8117 \\
Lateral        & 0.8782 & 0.7989 & 0.8182 & 0.8823 \\
Septal         & 0.8794 & 0.6763 & 0.7839 & 0.8927 \\
\hline
\textbf{Macro-AUC} & \textbf{0.8764} & \textbf{0.7759} & \textbf{0.8233} & \textbf{0.8817} \\
\hline
\end{longtable}

Using the MIMIC dataset for downstream evaluation (Table~\ref{tab:synthetic_eval}), the synthetic ECGs generated by our WearECG model achieved a macro-average AUROC of \textbf{0.8333}, closely approaching the original 12-lead configuration’s performance (0.8465) and substantially outperforming both the 3-lead (0.7837) and 1-lead (0.7545) setups. For critical cardiac conditions such as sinus bradycardia (AUROC = 0.9882), atrial fibrillation (AUROC = 0.9751), and premature ventricular complexes (AUROC = 0.9551), our model maintained high sensitivity and specificity, indicating effective preservation of key diagnostic features in the synthetic signals. 

To assess the generalizability of our model across datasets, we additionally evaluated downstream classification performance on the PTB-XL dataset. Consistent with the MIMIC results, WearECG-generated synthetic ECGs maintained competitive AUROC scores, demonstrating effective transfer of learned representations to a distinct clinical dataset.

\begin{longtable}{p{4.5cm}ccc}
\caption{AUROC comparison across different ECG lead configurations on PTBXL Dataset: WearECG (ours) and original 12-lead.} \\
\label{tab:synthetic_eval_PTBXL} \\
\toprule
Disease & WearECG (ours) & original 12-lead & 3-lead only\\
\midrule
\endfirsthead

\multicolumn{3}{c}{{\bfseries Table \thetable\ continued}} \\
\toprule
Disease & WearECG (ours) & original 12-lead & 3-lead only\\
\midrule
\endhead

\bottomrule
\endfoot

NORMAL & 0.8183 & 0.8468 & 0.8027 \\
SINUS RHYTHM & 0.8436 & 0.9393 & 0.7981 \\
SINUS BRADY & 0.9301 & 0.9881 & 0.9040\\
AFIB & 0.9883 & 0.9770 & 0.9917 \\
SINUS TACHY & 0.9907 & 0.9928 & 0.9844 \\
LAXIS DEV & 0.8311 & 0.8562 & 0.8102 \\
PVC & 0.9772 & 0.9391 & 0.9751 \\
RBBB & 0.9076 & 0.9588 & 0.8905 \\
LAE & 0.6621 & 0.8103 & 0.6848 \\
PAC & 0.9532 & 0.9570 & 0.8786 \\
PSVC & 0.8765 & 0.8852 & 0.8456 \\
LBBB & 0.9801 & 0.9848 & 0.8814 \\
LVH & 0.7283 & 0.7666 & 0.6673 \\
QT LONG & 0.8680 & 0.8224 & 0.8862 \\
AFLUTTER & 0.9589 & 0.9068 & 0.9906 \\
LAFB & 0.8384 & 0.9031 & 0.8152 \\
ECTOPIC ATR & 0.8337 & 0.8204 & 0.8170\\
RAE & 0.7511 & 0.8086 & 0.7243 \\
LPFB & 0.6734 & 0.8294 & 0.7857 \\
1ST AVB & 0.6857 & 0.9282 & 0.8375 \\
RVH & 0.8475 & 0.8927 & 0.8383 \\
ACUTE MI/STEMI & 0.6473 & 0.8734 & 0.5287\\
SVT & 0.8929 & 0.9540 & 0.8237\\
VT & 0.8641 & 0.8925 & 0.8264 \\

\textbf{Macro-AUC} & \textbf{0.8470} & \textbf{0.8931} & \textbf{0.8328}\\

\end{longtable}
 
\begin{longtable}{p{4.5cm}ccc}
\caption[Regional MI AUROC]{Regional MI classification on PTB-XL Dataset(AUROC)} \\[-1.5ex] % -1.5ex 可调整高度
\label{tab:MI_classification_PTBXL} \\
\toprule
Region & WearECG & Original 12-lead  & 3-lead only \\
\hline
\endhead
Anterior       & 0.5828 & 0.6465 & 0.6287 \\
Anterolateral  & 0.9112 & 0.9431 & 0.7473 \\
Anteroseptal   & 0.8756 & 0.9440 & 0.8400\\
Inferior       & 0.7701 & 0.8615 & 0.7374\\
Lateral        & 0.8460 & 0.9134 & 0.5594\\
Septal         & 0.7447 & 0.9464 & 0.6748 \\
\textbf{Macro-AUC} & \textbf{0.7885} & \textbf{0.8758} & \textbf{0.6979} \\
\bottomrule
\end{longtable}

Importantly, we further evaluated WearECG on regional myocardial infarction (MI) classification across six anatomical locations (anterior, anterolateral, anteroseptal, inferior, lateral, and septal). As shown in Table~\ref{tab:MI_classification}, our model consistently achieved high AUROC scores for all regions (macro-average AUROC = \textbf{0.8764}), demonstrating its ability to preserve spatially localized infarct patterns. This indicates that WearECG not only captures overall cardiac abnormalities but also retains the diagnostic fidelity necessary for identifying specific infarct locations, which is critical for accurate clinical assessment and intervention planning.  

In contrast, the 1-lead approach exhibited significantly degraded performance across most pathologies and infarct regions, highlighting its limited clinical utility. The 3-lead system, while better than 1-lead, still falls short of matching the fidelity of our reconstructed signals. These results emphasize the added value of our method in bridging the gap between minimal lead configurations and full 12-lead ECGs, making WearECG a promising tool for practical and accurate cardiac monitoring in resource-constrained settings.

Importantly, the successful transfer of knowledge from ECGFounder to synthetic ECGs underscores the generalizability and clinical value of our generation framework. This finding highlights the potential utility of synthetic ECGs in model pretraining, algorithm benchmarking, and data augmentation, especially in privacy-sensitive or low-resource settings where real ECG data are scarce or difficult to share.

\subsection*{WearECG can Improve the ECG Diagnosis of Cardiologists}

% \subsection*{WearECG can pass the Turing test}

Moreover, we also conducted a blinded Turing test involving three board-certified cardiologists. It also aims to evaluate the feature-level fidelity of the generated electrocardiogram (ECG) signals, we . A balanced set of 50 ECG samples—consisting of both real and generated signals—was randomly shuffled and presented under strictly blind conditions. The task required each expert to independently classify every sample as either “real” or “generated” based solely on waveform morphology and clinical plausibility. The resulting classification accuracies were 52\%, 44\%, and 44\%, respectively, which are statistically close to random guessing (i.e., 50\%). This suggests that the generated signals successfully mimic key morphological and temporal patterns found in authentic ECGs, such as waveform shapes, lead-to-lead coherence, and rhythm characteristics.

In the evaluation of model-assisted myocardial infarction (MI) recognition, we analyzed the same set of generated ECG samples from two complementary perspectives: a quantitative assessment using a confusion matrix and a qualitative examination through case studies.

For the confusion matrix evaluation (Figure X), the model-generated ECG samples, including both MI and normal cases, were assessed by physicians for MI detection and localization. The model achieved 26 true positives, 13 true negatives, 16 false positives, and 0 false negatives across 55 samples, corresponding to an accuracy of 70.9\%, sensitivity of 100\%, specificity of 44.8\%, precision of 61.9\%, and an F1-score of 76.5\%.The results highlight that the model can effectively assist physicians in identifying MI, achieving high sensitivity while maintaining moderate specificity.

\begin{figure}[H]
\centering
\includegraphics[width=0.6\textwidth]{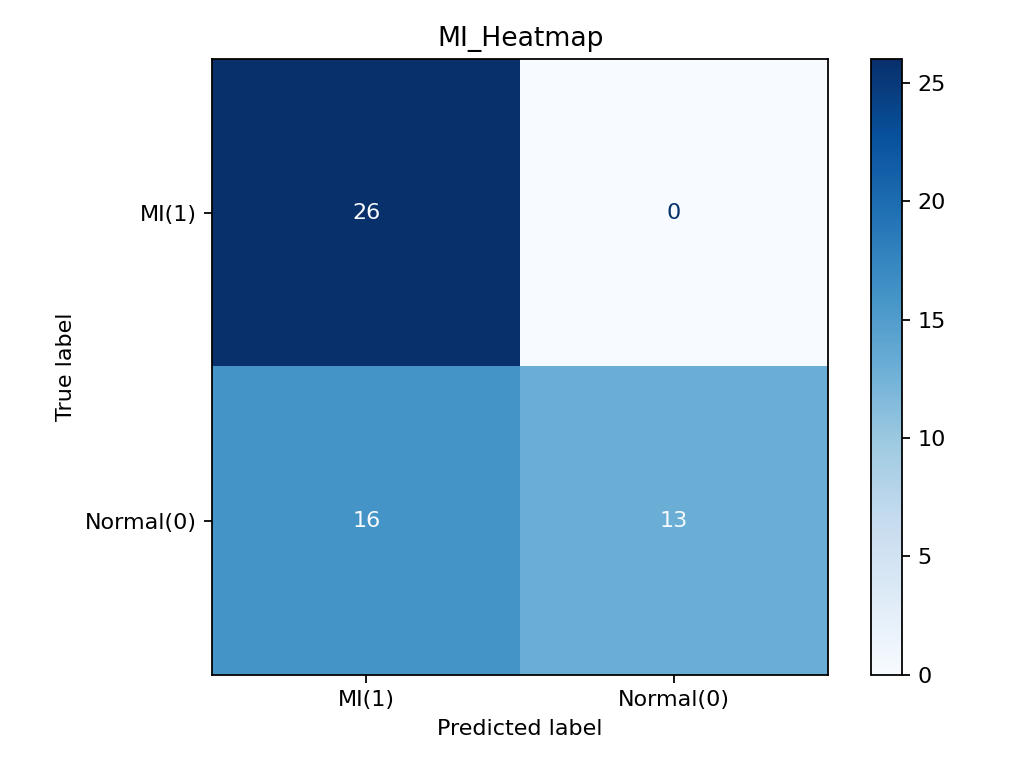} % 替换为实际图片路径
\caption{Confusion matrix for model-assisted myocardial infarction (MI) detection.}
\label{fig:confmat_MI}
\end{figure}

In the case study analysis, we selected representative samples to evaluate the effectiveness of model-generated ECGs in assisting physicians with myocardial infarction (MI) localization. Accurate localization of infarct regions is critical for guiding timely and appropriate clinical interventions, including treatment planning and risk assessment. 

\begin{table}[H]
\centering
\small
\caption{Representative case study samples for model-assisted MI localization.}
\begin{tabular}{|c|c|c|c|}
\hline
\textbf{Physician 1} & \textbf{Physician 2} & \textbf{Physician 3} & \textbf{Model Prediction} \\
\hline
Septal & Anteroseptal & Anteroseptal & Anteroseptal \\
Inferior/Septal & Inferior & Inferior & Inferior \\
\hline
\end{tabular}
\label{tab:case_study_red}
\end{table}

These results indicate that the model-generated ECGs closely align with physician assessments, effectively assisting in MI localization and providing useful reference for clinical interpretation. By accurately reconstructing ECG signals, the model offers a valuable tool for detecting and localizing myocardial infarction, which is critical for timely clinical decision-making, risk assessment, and treatment planning. This demonstrates that reconstructed ECGs can play an important role in supporting cardiologists in diagnosing heart conditions, particularly in identifying infarct regions, and may help reduce the likelihood of misdiagnosis in clinical practice. Additional cases are provided in the appendix for further reference.

\section*{DISCUSSION}

%% Introduction
In this study, we present a novel ECG signal generation and reconstruction framework aimed at synthesizing 12-lead ECGs from only three input leads. This approach effectively reduces data acquisition costs while maintaining high-fidelity signal reconstruction and diagnostic utility. Leveraging a variational autoencoder (VAE)-based backbone and integrating downstream classification heads for clinical diagnosis tasks, our pipeline offers a comprehensive solution for efficient and accurate ECG signal generation. Experimental results validate the robustness of the generated signals in both waveform-level metrics and downstream diagnostic performance, demonstrating the clinical promise of low-lead ECG reconstruction.

%% contributions
Our main contribution lies in proposing a systematic pipeline for 3-lead to 12-lead ECG generation, evaluated not only by signal similarity metrics (e.g., FID, MSE) but also through downstream diagnostic tasks using a frozen expert model (ECGFounder). This dual-evaluation strategy provides insight into both the signal fidelity and clinical validity of the generated ECGs. Importantly, we achieve effective reconstruction and myocardial infarction localization from 3-lead inputs on portable devices, reaching diagnostic performance comparable to clinical reference standards on real-world datasets. Additionally, we demonstrate that the generated signals retain sufficient pathological information to support reliable myocardial infarction classification. Our design also allows modular adaptation, making it feasible to incorporate different generation backbones or clinical models in future studies.

%% social impact
The proposed framework holds significant potential in resource-constrained medical scenarios, such as remote monitoring, wearable devices, and low-cost screening systems. By leveraging AI models to reconstruct high-quality 12-lead ECG signals from only 3-lead inputs, it not only preserves the essential diagnostic information but also enables effective localization of myocardial infarction. This breakthrough facilitates portable and accessible MI detection, which is crucial for reducing cases of missed or delayed medical intervention, especially in underserved or remote regions. Moreover, the system can be integrated into the clinical pathway spanning chest pain centers, pre-hospital emergency response, and community-based diagnostic services. This integration ensures early identification of critical cardiac events and provides robust auxiliary support for triage and referral decisions in environments where advanced ECG equipment or specialized cardiologists are unavailable. Additionally, by improving the completeness and reliability of diagnostic data derived from limited leads, this approach enhances clinical coverage in challenging scenarios characterized by patient movement, noise interference, or emergency conditions where acquiring full 12-lead ECGs is impractical.

%% Limitations
Despite promising results, our current framework still faces several important limitations. First, although the model demonstrates strong performance on the MIMIC dataset, this dataset mainly represents a specific patient population and device setting. Therefore, the generalization ability of the model to other populations with different ethnic, demographic, or clinical characteristics, as well as to ECG data acquired from different devices or clinical environments, remains unverified. This gap limits the immediate applicability of the model in diverse real-world scenarios and calls for further external validation on heterogeneous datasets.The current framework operates solely on raw ECG waveforms without leveraging complementary clinical diagnostic information, such as cardiologist textual reports or structured clinical notes. This lack of multimodal semantic integration means that the model cannot fully exploit the rich contextual and interpretative information embedded in clinical texts, which could be valuable for improving both the fidelity and diagnostic relevance of the generated signals. While ECGFounder provides a powerful and pretrained diagnostic backbone that offers objective quantitative evaluation of generated ECGs, it remains an algorithmic proxy and cannot replace expert cardiologist assessment. The subtle nuances and complex judgment calls made by experienced clinicians during ECG interpretation involve contextual knowledge and clinical experience that current automated models cannot yet replicate. Therefore, extensive clinical validation involving human experts remains essential to confirm the safety, reliability, and clinical utility of the generated ECG signals before deployment in real medical practice.

%% future work
Future work will focus on three main directions. First, we aim to rigorously validate the generalizability of our model across diverse datasets and real-world clinical ECG recordings that exhibit varying patient demographics, device types, noise levels, and lead configurations. This will help ensure robustness and applicability of the model in heterogeneous clinical settings. Moreover, we plan to enhance our downstream evaluation framework by including a wider range of cardiovascular conditions beyond myocardial infarction, such as arrhythmias and conduction abnormalities, to comprehensively assess the diagnostic value and clinical relevance of the generated ECG signals. Third, recognizing the current limitation of lacking semantic clinical context, we intend to integrate multimodal information, such as clinical diagnostic text reports and structured medical records, into the generation process. This integration will provide richer semantic constraints and improve both interpretability and diagnostic utility. Additionally, we will explore the incorporation of physiological priors and anatomical constraints—like heart vector models or lead field theory—into the model architecture to enhance biological plausibility and signal realism. In the long term, we envision adapting this framework for real-time implementation in portable or wearable ECG devices, enabling low-cost, accessible cardiovascular monitoring and early diagnostic support in resource-limited and ambulatory environments.

\newpage

\section*{METHODS}

%%%  The methods (also/formerly called "experimental 
%%%  procedures") should appear 
%%%  immediately after the discussion. Subheadings 
%%%  may be customized. Please consult your handling 
%%%  editor if you have questions about what content 
%%%  should appear here. 

\subsection*{Problem Definition}
The electrocardiogram (ECG) is a non-invasive tool that records the heart's electrical activity over time, characterized by waveforms such as the P-wave, QRS complex, and T-wave, which correspond to distinct phases of the cardiac cycle. The standard 12-lead ECG, derived from 10 electrodes placed on the limbs and chest (as detailed in Table~\ref{tab:ecg_leads}), provides a spatially comprehensive view of cardiac function and is widely used for diagnosing various cardiovascular conditions.

\begin{table}[H]
\centering
\caption{Electrode configuration in the standard 12-lead ECG system.}
\label{tab:ecg_leads}
\small
\begin{tabular}{ll}
\toprule
\textbf{Lead} & \textbf{Electrode Position} \\
\midrule
I    & Left Arm, Right Arm \\
II   & Left Foot, Right Arm \\
III  & Left Foot, Left Arm \\
aVR  & Right Arm \\
aVL  & Left Arm \\
aVF  & Left Foot \\
V1   & 4th intercostal space, right sternal border \\
V2   & 4th intercostal space, left sternal border \\
V3   & Midpoint between V2 and V4 \\
V4   & 5th intercostal space, midclavicular line \\
V5   & Lateral to V4, left midaxillary line \\
V6   & Lateral to V5, left midaxillary line \\
\bottomrule
\end{tabular}
\end{table}

We formulate 12-lead ECG reconstruction as a generative task, where the input contains only partial information from three selected leads. Given a segment $x \in \mathbb{R}^{L \times T}$ representing an ECG with $L=12$ leads over $T$ time steps (e.g., $T = 1000$ for 2 seconds at 500Hz), we simulate 3-lead input by masking out 9 leads:

\begin{equation}
    x_{masked} = x \odot (1 - M)
\end{equation}

Here, $M \in \{0,1\}^{L \times T}$ is a binary mask with zeros at the retained leads II, V1, V5 and ones elsewhere. The model is trained to reconstruct the full 12-lead ECG $y \in \mathbb{R}^{12 \times T}$ from $x_{masked}$.

We adopt a standard Variational Autoencoder (VAE) framework. The encoder maps the masked ECG to a latent distribution $q(z|x_{masked})$, from which we sample a latent vector $z$. The decoder then reconstructs the full ECG $\hat{y} = f_\theta(z)$.

The model is optimized using a combination of:
\begin{itemize}
  \item Mean squared error between $\hat{y}$ and $y$
  \item KL divergence between the approximate posterior and prior
  \item Optional perceptual loss based on a pretrained ECG encoder
\end{itemize}

To assess robustness and clinical applicability, we apply comparative experiments with multiple lead configurations, including:
\begin{itemize}
  \item Single-lead input (e.g., lead I only)
  \item Three-lead input (e.g., I–II–V3 and II–V1–V5)
\end{itemize}

This setup enables the model to learn to infer missing leads from partial observations, mimicking clinical scenarios where limited-lead monitoring is available. What's more, it reflects realistic scenarios in low-resource or wearable monitoring environments.

\subsection*{WearECG Overview}\label{overview}

We propose a two-stage framework to reconstruct 12-lead ECG signals from 3-lead inputs and enable multi-label disease classification. The first stage adopts an enhanced Variational Autoencoder (VAE) architecture \cite{kingma2022autoencodingvariationalbayes}, which extends the basic VAE structure by introducing several critical improvements for ECG signal modeling: (1) deep residual convolutional blocks that capture richer temporal representations; (2) multi-head self-attention modules to model long-range dependencies between leads; and (3) a hierarchical encoder-decoder design with progressive downsampling and upsampling to aggregate global context efficiently. This symmetric encoder-decoder backbone leverages structured latent representations through reparameterization to generate full-lead ECGs from partial inputs. Additionally, a perceptual loss is optionally employed during training to better preserve clinically meaningful features in the reconstructed signals, enhancing their diagnostic value.

In the second stage, we fine-tune ECGFounder \cite{li2024ecgfounder}, a pre-trained ECG classification model, by attaching a lightweight classification head for downstream diagnosis tasks such as myocardial infarction, conduction blocks, and arrhythmias, formulated as multi-label prediction. The entire pipeline is modular and can be trained end-to-end or in a decoupled manner, demonstrating robustness and clinical relevance across various lead configurations.
\begin{figure}[H]
\centering
\includegraphics[width=0.95\textwidth]{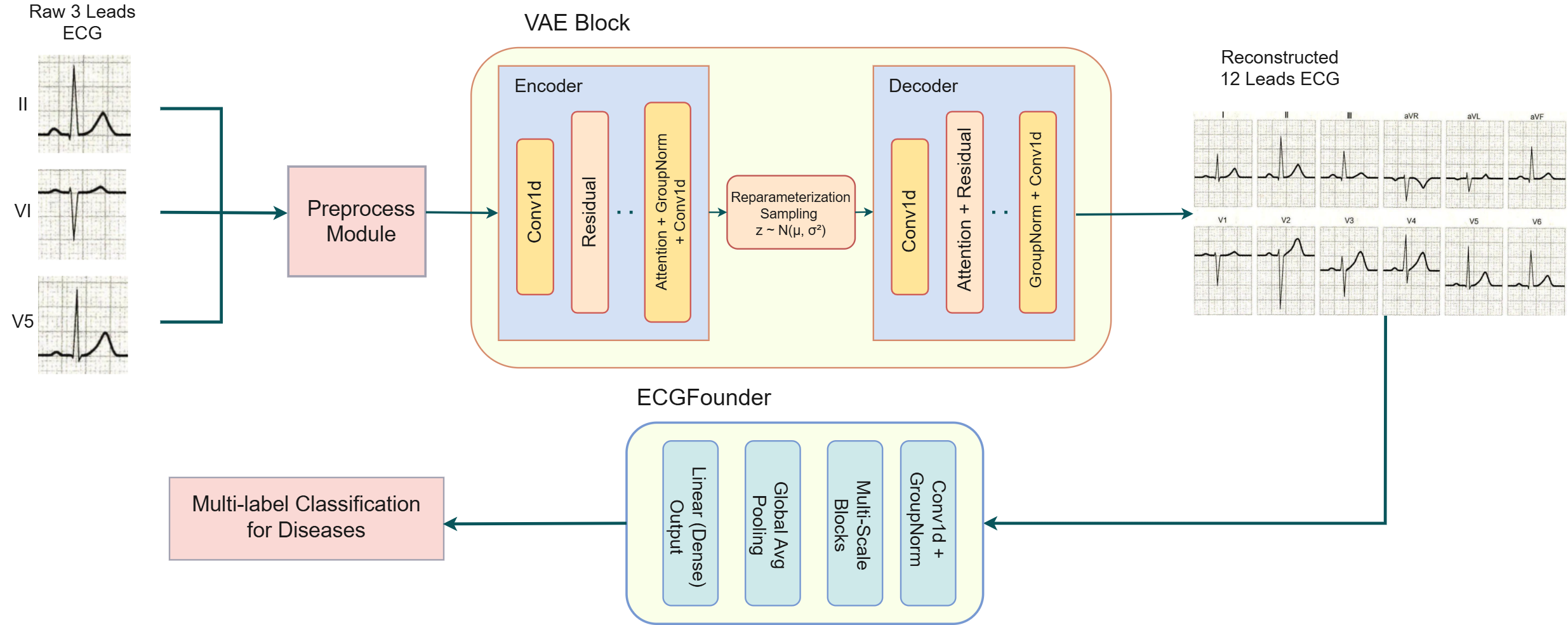}
\caption{Framework Overview}
\label{framework-fig}
\end{figure}

\subsection*{Dataset and Data Preprocessing}

The raw 12-lead ECG signals underwent a multi-step preprocessing pipeline to ensure data quality and uniformity for model training. First, the leads were reordered to a standardized sequence to maintain consistency across datasets by mapping their original order to a target reference layout. Next, all signals were resampled to a fixed frequency of 500 Hz using Fourier-based interpolation, which preserved signal integrity while adjusting the time series length. Missing values (NaNs) within the signals were replaced by averaging neighboring data points within a defined window, effectively mitigating missing data without causing significant bias. 

Although optional in this study, per-lead Z-score normalization was applied to align data distributions across samples and aid model convergence. To handle the large dataset efficiently, records were processed in batches, each undergoing NaN replacement, resampling, and optional normalization before saving. Finally, to prevent data leakage, train-test splits were created by grouping records according to \textit{subject\_id}, ensuring that all records from a given subject appear exclusively in either the training or testing set.
\subsection*{WearECG Encoder: Learning Latent Structures from Limited Leads}
The encoder is constructed using a multi-layer one-dimensional convolutional neural network (1D CNN) tailored to process 12-lead electrocardiogram (ECG) signals. The input signal \(\mathbf{x} \in \mathbb{R}^{T \times 12}\), where \(T\) is the temporal length, is initially projected to a higher dimensional feature space through a convolutional layer, increasing channels from 12 to 128. The network then applies residual blocks and downsampling layers to extract hierarchical features, reducing temporal resolution while increasing channel depth to 512, resulting in a compressed representation \(\mathbf{h} \in \mathbb{R}^{T' \times C}\) with \(T' < T\).

Two separate convolutional layers predict the parameters of the approximate posterior distribution over the latent space:

\begin{equation}
\boldsymbol{\mu} = f_{\mu}(\mathbf{h}), \quad \log \boldsymbol{\sigma}^2 = f_{\sigma}(\mathbf{h}),
\end{equation}

where \(\boldsymbol{\mu}, \log \boldsymbol{\sigma}^2 \in \mathbb{R}^{T' \times d}\), and \(d\) is the latent dimension (here \(d=4\)).

The latent variable \(\mathbf{z}\) is then sampled using the reparameterization trick:

\begin{equation}
\mathbf{z} = \boldsymbol{\mu} + \boldsymbol{\sigma} \odot \boldsymbol{\epsilon}, \quad \boldsymbol{\epsilon} \sim \mathcal{N}(\mathbf{0}, \mathbf{I}),
\end{equation}

where \(\odot\) denotes element-wise multiplication, ensuring differentiability of the sampling process for backpropagation.

\subsection*{WearECG Decoder: Reconstructing Twelve-Lead Signals from Latent Space}
The decoder receives the latent representation \(\mathbf{z} \in \mathbb{R}^{T' \times d}\) and aims to reconstruct the original ECG signal \(\hat{\mathbf{x}} \in \mathbb{R}^{T \times 12}\). It employs a series of residual blocks integrated with multi-head self-attention mechanisms, which model long-range temporal dependencies. The decoder progressively upsamples the temporal dimension from \(T'\) back to \(T\) while reducing the channel depth from \(d\) back to 12.

Formally, the decoding process can be described as:

\begin{equation}
\hat{\mathbf{x}} = g_{\theta}(\mathbf{z}),
\end{equation}

where \(g_{\theta}(\cdot)\) represents the decoder network parameterized by \(\theta\), incorporating convolutional, residual, attention, and upsampling layers.

The model is trained under the variational autoencoder framework by optimizing the evidence lower bound (ELBO):

\begin{equation}
\mathcal{L}(\theta, \phi; \mathbf{x}) = \mathbb{E}_{q_{\phi}(\mathbf{z}|\mathbf{x})} \big[\log p_{\theta}(\mathbf{x}|\mathbf{z})\big] - \mathrm{KL}\big(q_{\phi}(\mathbf{z}|\mathbf{x}) \,||\, p(\mathbf{z})\big),
\end{equation}

where

\begin{equation}
q_{\phi}(\mathbf{z}|\mathbf{x}) = \mathcal{N}(\mathbf{z}; \boldsymbol{\mu}, \mathrm{diag}(\boldsymbol{\sigma}^2)), \quad p(\mathbf{z}) = \mathcal{N}(\mathbf{0}, \mathbf{I}),
\end{equation}

with \(q_{\phi}\) the encoder’s approximate posterior and \(p(\mathbf{z})\) the prior.

The reconstruction loss is computed as the mean squared error (MSE):

\begin{equation}
\mathcal{L}_{\text{recon}} = \frac{1}{N}\sum_{i=1}^N \| \mathbf{x}_i - \hat{\mathbf{x}}_i \|^2,
\end{equation}

and the KL divergence term has a closed-form expression:

\begin{equation}
\mathrm{KL} = -\frac{1}{2} \sum_{j=1}^d \left( 1 + \log \sigma_j^2 - \mu_j^2 - \sigma_j^2 \right).
\end{equation}

The total loss is thus:

\begin{equation}
\mathcal{L} = \mathcal{L}_{\text{recon}} + \beta \cdot \mathrm{KL},
\end{equation}

where \(\beta\) is a hyperparameter weighting the KL term (e.g., \(10^{-4}\)).

Optimization is performed using the AdamW optimizer, with parameters updated by

\begin{equation}
\theta, \phi \leftarrow \theta, \phi - \eta \nabla_{\theta, \phi} \mathcal{L},
\end{equation}

where \(\eta\) is the learning rate scheduled by a OneCycle learning rate policy to facilitate convergence.

To avoid data leakage, the dataset is split by subject according to their \textit{subject\_id}, ensuring that all samples from a single subject belong exclusively to either the training or testing sets.

\subsection*{Fine-tuned ECGFounder}
In the downstream classification stage, we fine-tune ECGFounder, a deep neural network pretrained on large-scale 12-lead ECG datasets.This study addresses a multi-label classification task on 12-lead ECG signals, targeting approximately 40 cardiac disease categories, including myocardial infarction\cite{liu2020classification}, various conduction blocks\cite{li2021classification}, and arrhythmias. Labels are represented as multi-hot vectors, enabling simultaneous detection of multiple coexisting conditions. The task demands robust generalization and fine-grained feature extraction from the model.

ECGFounder employs a modular 1D convolutional architecture tailored for multi-lead ECG signals. The backbone network consists of multiple residual bottleneck blocks, each containing sequential 1×1, k×1, and 1×1 convolutions, combined with batch normalization and Swish activation functions. To maintain temporal resolution and reduce information loss, custom SAME-padding convolution and max-pooling layers are applied. The network progressively downsamples the input signal through multiple stages, aggregating hierarchical features while incorporating squeeze-and-excitation modules for channel-wise attention.

Prior to model input, each ECG lead is standardized using Z-score normalization \cite{shukla2023arrhythmia} to mitigate offset and amplitude scaling issues.During fine-tuning, the pretrained weights are loaded, and the backbone parameters are frozen to retain learned representations. A lightweight classification head—a fully connected linear layer—is appended to the extracted deep features for multi-label prediction of cardiovascular conditions such as myocardial infarction, conduction blocks, and arrhythmias. Input ECG signals are standardized per lead by Z-score normalization before feeding into the model. Training uses a multi-label binary cross-entropy loss, optimized with Adam, with hyperparameters empirically selected for convergence.

The entire pipeline demonstrates robustness and clinical relevance across various lead configurations, enabling effective downstream diagnosis from reconstructed or raw ECG signals.

\subsection*{Training Setup}

The model was trained on a single NVIDIA RTX 3090 GPU using the PyTorch framework. We utilized an efficient data loading pipeline that shuffled the training data and processed it in batches of size 16. To optimize the model parameters, the AdamW optimizer was employed with an initial learning rate of \(1 \times 10^{-5}\). A OneCycle learning rate scheduler was applied, with a maximum learning rate of \(5 \times 10^{-5}\) and a warm-up period corresponding to 20\% of the total training steps, which facilitated faster and more stable convergence.

Training was conducted for 10 epochs. The loss function combined a mean squared reconstruction loss and a Kullback–Leibler divergence (KLD) regularization term, where the KLD weight was set to \(1 \times 10^{-4}\) to balance reconstruction fidelity and latent space regularization. To reduce GPU memory consumption and accelerate training without compromising performance, mixed precision training was enabled via PyTorch's Automatic Mixed Precision (AMP) feature.

To prevent data leakage and ensure the robustness of model evaluation, the dataset was split by \textit{subject\_id}, such that all samples belonging to the same subject were contained exclusively in either the training or validation sets. Data preprocessing, including normalization and consistent lead reordering, was performed during data loading to maintain input uniformity.

Model checkpoints were saved at the end of each epoch to enable recovery and facilitate model selection. Although an early stopping mechanism was configured, it was not triggered as training proceeded stably without overfitting.

\newpage

%%%  The following components should appear after the 
%%%  methods. 
%%%  For journals using STAR Methods, these components 
%%%  should appear immediately after the discussion 
%%%  (after any "limitations" or "conclusions" subsection 
%%%  within the discussion).

\section*{RESOURCE AVAILABILITY}

% The code of our method and evaluations is publicly available at \url{https://github.com/Raiiyf/DiffuSETS_Exp}.
The code used in this study is not publicly available at the moment. It will be made available on GitHub upon publication. Data used in this study (MIMIC-III ECG) are available under standard access agreements from PhysioNet.

\section*{ACKNOWLEDGMENTS}

%%%  Use this section to acknowledge contributions 
%%%  from non-authors and list funding sources, 
%%%  including grant numbers.

This work is supported by the National Natural Science Foundation of China (62102008, 62172018), CCF-Tencent Rhino-Bird Open Research Fund (CCF-Tencent RAGR20250108), CCF-Zhipu Large Model Innovation Fund (CCF-Zhipu202414), PKU-OPPO Fund (BO202301, BO202503), Research Project of Peking University in the State Key Laboratory of Vascular Homeostasis and Remodeling (2025-SKLVHR-YCTS-02).

% \section*{AUTHOR CONTRIBUTIONS}

% %%%  This component is required for most research papers. 
% %%%  Mention each individual author with a statement 
% %%%  outlining the contribution of each author to the work.

% Conceptualization, S.C.P. and S.Y.W.; methodology, A.B., S.C.P., and S.Y.W.; investigation, M.E., A.N.V., N.A.V., S.C.P., and S.Y.W.; writing-–original draft, S.C.P. and S.Y.W.; writing-–review \& editing, S.C.P. and S.Y.W.; funding acquisition, S.C.P. and S.Y.W.; resources, M.E.V and C.K.B.; supervision, A.B., N.L.W., and A.A.D.

\section*{DECLARATION OF INTERESTS}

%%%  This component is required for all articles, even 
%%%  if the authors have no competing interests; if 
%%%  this is the case, insert "The authors declare no 
%%%  competing interests." Please refer to the 
%%%  declaration of interests policy: 
%%%  https://www.cell.com/declaration-of-interests

The authors declare no competing interests.

\newpage

%%%  REFERENCES: As of 2023, all Cell Press journals 
%%%  use Numbered (AMA) style. We recommend placing 
%%%  your references in the included "references.bib" 
%%%  file.

\bibliography{reference}

\begin{thebibliography}{30}
\providecommand{\natexlab}[1]{#1}
\providecommand{\url}[1]{\texttt{#1}}
\providecommand{\href}[2]{#2}
\providecommand{\path}[1]{#1}
\providecommand{\DOIprefix}{doi: }
\providecommand{\ArXivprefix}{arXiv: }
\providecommand{\URLprefix}{URL: }
\providecommand{\Pubmedprefix}{pmid: }
\providecommand{\doi}[1]{\href{http://dx.doi.org/#1}{\path{#1}}}
\providecommand{\Pubmed}[1]{\href{pmid:#1}{\path{#1}}}
\providecommand{\BIBand}{and}
\providecommand{\bibinfo}[2]{#2}
\ifx\xfnm\undefined \def\xfnm[#1]{\unskip,\space#1}\fi
\makeatletter\def\@biblabel#1{#1.}\makeatother
%Type = Article
\bibitem[{Nabel(2003)}]{Nabel2003}
\bibinfo{author}{Nabel, E.G.} (\bibinfo{year}{2003}). \bibinfo{title}{Cardiovascular disease}.
\newblock \bibinfo{journal}{New England Journal of Medicine} \emph{\bibinfo{volume}{349}}, \bibinfo{pages}{60--72}.
%Type = Article
\bibitem[{Roth et~al.(2018)}]{Roth2018}
\bibinfo{author}{Roth, G.A.} et~al. (\bibinfo{year}{2018}). \bibinfo{title}{Global, regional, and national age-sex-specific mortality for 282 causes of death in 195 countries and territories, 1980--2017: a systematic analysis for the global burden of disease study 2017}.
\newblock \bibinfo{journal}{Lancet} \emph{\bibinfo{volume}{392}}, \bibinfo{pages}{1736--1788}.
%Type = Article
\bibitem[{Turakhia et~al.(2013)}]{Turakhia2013}
\bibinfo{author}{Turakhia, M.P.} et~al. (\bibinfo{year}{2013}). \bibinfo{title}{Diagnostic utility of a novel leadless arrhythmia monitoring device}.
\newblock \bibinfo{journal}{American Journal of Cardiology} \emph{\bibinfo{volume}{112}}, \bibinfo{pages}{520--524}.
%Type = Article
\bibitem[{Lai et~al.(2020)Lai, Bu, Su, Zhang and Ma}]{Lai2020}
\bibinfo{author}{Lai, D.}, \bibinfo{author}{Bu, Y.}, \bibinfo{author}{Su, Y.}, \bibinfo{author}{Zhang, X.}, and \bibinfo{author}{Ma, C.S.} (\bibinfo{year}{2020}). \bibinfo{title}{Non-standardized patch-based ecg lead together with deep learning based algorithm for automatic screening of atrial fibrillation}.
\newblock \bibinfo{journal}{IEEE Journal of Biomedical and Health Informatics} \emph{\bibinfo{volume}{24}}, \bibinfo{pages}{1569--1578}.
%Type = Article
\bibitem[{Liu et~al.(2021)}]{Liu2021}
\bibinfo{author}{Liu, C.M.} et~al. (\bibinfo{year}{2021}). \bibinfo{title}{Enhanced detection of cardiac arrhythmias utilizing 14-day continuous ecg patch monitoring}.
\newblock \bibinfo{journal}{International Journal of Cardiology} \emph{\bibinfo{volume}{332}}, \bibinfo{pages}{78--84}.
%Type = Article
\bibitem[{Tison et~al.(2018)}]{Tison2018}
\bibinfo{author}{Tison, G.H.} et~al. (\bibinfo{year}{2018}). \bibinfo{title}{Passive detection of atrial fibrillation using a commercially available smartwatch}.
\newblock \bibinfo{journal}{JAMA Cardiology} \emph{\bibinfo{volume}{3}}, \bibinfo{pages}{409--416}.
%Type = Article
\bibitem[{Bumgarner et~al.(2018)}]{Bumgarner2018}
\bibinfo{author}{Bumgarner, J.M.} et~al. (\bibinfo{year}{2018}). \bibinfo{title}{Smartwatch algorithm for automated detection of atrial fibrillation}.
\newblock \bibinfo{journal}{Journal of the American College of Cardiology} \emph{\bibinfo{volume}{71}}, \bibinfo{pages}{2381--2388}.
%Type = Article
\bibitem[{Perez et~al.(2019)}]{Perez2019}
\bibinfo{author}{Perez, M.V.} et~al. (\bibinfo{year}{2019}). \bibinfo{title}{Large-scale assessment of a smartwatch to identify atrial fibrillation}.
\newblock \bibinfo{journal}{New England Journal of Medicine} \emph{\bibinfo{volume}{381}}, \bibinfo{pages}{1909--1917}.
%Type = Article
\bibitem[{Rachim and Chung(2016)}]{Rachim2016}
\bibinfo{author}{Rachim, V.P.}, and \bibinfo{author}{Chung, W.Y.} (\bibinfo{year}{2016}). \bibinfo{title}{Wearable noncontact armband for mobile ecg monitoring system}.
\newblock \bibinfo{journal}{IEEE Transactions on Biomedical Circuits and Systems} \emph{\bibinfo{volume}{10}}, \bibinfo{pages}{1112--1118}.
%Type = Article
\bibitem[{Li et~al.(2021{\natexlab{a}})}]{Li2021}
\bibinfo{author}{Li, B.} et~al. (\bibinfo{year}{2021}{\natexlab{a}}). \bibinfo{title}{Influence of armband form factors on wearable ecg monitoring performance}.
\newblock \bibinfo{journal}{IEEE Sensors Journal} \emph{\bibinfo{volume}{21}}, \bibinfo{pages}{11046--11060}.
%Type = Article
\bibitem[{L{\'a}zaro et~al.(2020)}]{Lazaro2020}
\bibinfo{author}{L{\'a}zaro, J.} et~al. (\bibinfo{year}{2020}). \bibinfo{title}{Wearable armband device for daily life electrocardiogram monitoring}.
\newblock \bibinfo{journal}{IEEE Transactions on Biomedical Engineering} \emph{\bibinfo{volume}{67}}, \bibinfo{pages}{3464--3473}.
%Type = Article
\bibitem[{Nelwan et~al.(2000)Nelwan, Kors and Meij}]{nelwan2000minimal}
\bibinfo{author}{Nelwan, S.}, \bibinfo{author}{Kors, J.}, and \bibinfo{author}{Meij, S.} (\bibinfo{year}{2000}). \bibinfo{title}{Minimal lead sets for reconstruction of 12-lead electrocardiograms}.
\newblock \bibinfo{journal}{Journal of Electrocardiology} \emph{\bibinfo{volume}{33}}, \bibinfo{pages}{163--166}.
%Type = Article
\bibitem[{Maheshwari et~al.(2016)Maheshwari, Acharyya, Schiariti and Puddu}]{maheshwari2016frank}
\bibinfo{author}{Maheshwari, S.}, \bibinfo{author}{Acharyya, A.}, \bibinfo{author}{Schiariti, M.}, and \bibinfo{author}{Puddu, P.} (\bibinfo{year}{2016}). \bibinfo{title}{Frank vectorcardiographic system from standard 12 lead ecg: An effort to enhance cardiovascular diagnosis}.
\newblock \bibinfo{journal}{Journal of Electrocardiology} \emph{\bibinfo{volume}{49}}, \bibinfo{pages}{231--242}.
%Type = Article
\bibitem[{Lee et~al.(2016)Lee, Kim and Kim}]{lee2016state}
\bibinfo{author}{Lee, J.}, \bibinfo{author}{Kim, M.}, and \bibinfo{author}{Kim, J.} (\bibinfo{year}{2016}). \bibinfo{title}{Reconstruction of precordial lead electrocardiogram from limb leads using the state-space model}.
\newblock \bibinfo{journal}{IEEE Journal of Biomedical and Health Informatics} \emph{\bibinfo{volume}{20}}, \bibinfo{pages}{818--828}.
%Type = Article
\bibitem[{Zhu et~al.(2018)Zhu, Pan, Cheng and Huan}]{zhu2018lightweight}
\bibinfo{author}{Zhu, H.}, \bibinfo{author}{Pan, Y.}, \bibinfo{author}{Cheng, K.}, and \bibinfo{author}{Huan, R.} (\bibinfo{year}{2018}). \bibinfo{title}{A lightweight piecewise linear synthesis method for standard 12-lead ecg signals based on adaptive region segmentation}.
\newblock \bibinfo{journal}{PLoS One} \emph{\bibinfo{volume}{13}}, \bibinfo{pages}{e0206170}.
%Type = Article
\bibitem[{Tsouri and Ostertag(2014)}]{tsouri2014patient}
\bibinfo{author}{Tsouri, G.}, and \bibinfo{author}{Ostertag, M.} (\bibinfo{year}{2014}). \bibinfo{title}{Patient-specific 12-lead ecg reconstruction from sparse electrodes using independent component analysis}.
\newblock \bibinfo{journal}{IEEE Journal of Biomedical and Health Informatics} \emph{\bibinfo{volume}{18}}, \bibinfo{pages}{476--482}.
%Type = Inproceedings
\bibitem[{Nejedly et~al.(2021)Nejedly, Ivora, Smisek, Viscor, Koscova, Jurak and Plesinger}]{nejedly2021}
\bibinfo{author}{Nejedly, P.}, \bibinfo{author}{Ivora, A.}, \bibinfo{author}{Smisek, R.}, \bibinfo{author}{Viscor, I.}, \bibinfo{author}{Koscova, Z.}, \bibinfo{author}{Jurak, P.}, and \bibinfo{author}{Plesinger, F.} (\bibinfo{year}{2021}). \bibinfo{title}{Classification of ecg using ensemble of residual cnns with attention mechanism}.
\newblock In \bibinfo{booktitle}{Computing in Cardiology (CinC)}. pp. \bibinfo{pages}{1--4}.
\newblock \URLprefix \url{https://doi.org/10.23919/CinC53138.2021.9662723}. \DOIprefix\doi{10.23919/CinC53138.2021.9662723}.
%Type = Inproceedings
\bibitem[{Gundlapalle and Acharyya(2022)}]{gundlapalle2022}
\bibinfo{author}{Gundlapalle, V.}, and \bibinfo{author}{Acharyya, A.} (\bibinfo{year}{2022}). \bibinfo{title}{A novel single lead to 12-lead ecg reconstruction methodology using convolutional neural networks and lstm}.
\newblock In \bibinfo{booktitle}{2022 IEEE 13th Latin America Symposium on Circuits and System (LASCAS)}. pp. \bibinfo{pages}{01--04}.
\newblock \DOIprefix\doi{10.1109/LASCAS53948.2022.9789045}.
%Type = Inproceedings
\bibitem[{Garg et~al.(2023)Garg, Venkataramani and Priyakumar}]{garg2023}
\bibinfo{author}{Garg, A.}, \bibinfo{author}{Venkataramani, V.V.}, and \bibinfo{author}{Priyakumar, U.D.} (\bibinfo{year}{2023}). \bibinfo{title}{Single-lead to multi-lead electrocardiogram reconstruction using a modified attention u-net framework}.
\newblock In \bibinfo{booktitle}{2023 International Joint Conference on Neural Networks (IJCNN)}. pp. \bibinfo{pages}{1--8}.
\newblock \DOIprefix\doi{10.1109/IJCNN54540.2023.10191213}.
%Type = Article
\bibitem[{Chen et~al.(2024)Chen, Wu, Liu and Hong}]{chen2024}
\bibinfo{author}{Chen, J.}, \bibinfo{author}{Wu, W.}, \bibinfo{author}{Liu, T.}, and \bibinfo{author}{Hong, S.} (\bibinfo{year}{2024}). \bibinfo{title}{Multi-channel masked autoencoder and comprehensive evaluations for reconstructing 12-lead ecg from arbitrary single-lead ecg}.
\newblock \bibinfo{journal}{Preprint or Conference Name}.
\newblock \bibinfo{note}{Available upon request or add proper publication details}.
%Type = Article
\bibitem[{Lee et~al.(2020)Lee, Lee, Lee, Kim and Kim}]{lee2020synthesis}
\bibinfo{author}{Lee, J.}, \bibinfo{author}{Lee, Y.}, \bibinfo{author}{Lee, G.H.}, \bibinfo{author}{Kim, K.S.}, and \bibinfo{author}{Kim, K.S.} (\bibinfo{year}{2020}). \bibinfo{title}{Synthesis of electrocardiogram v-lead signals from limb-lead measurement using r-peak aligned generative adversarial network}.
\newblock \bibinfo{journal}{IEEE Journal of Biomedical and Health Informatics} \emph{\bibinfo{volume}{24}}, \bibinfo{pages}{1265--1275}.
%Type = Article
\bibitem[{Seo et~al.(2022)Seo, Lee, Kim, Kim and Cho}]{seo2022multiple}
\bibinfo{author}{Seo, J.H.}, \bibinfo{author}{Lee, I.H.}, \bibinfo{author}{Kim, S.W.}, \bibinfo{author}{Kim, S.H.}, and \bibinfo{author}{Cho, S.J.} (\bibinfo{year}{2022}). \bibinfo{title}{Multiple electrocardiogram generator with single-lead electrocardiogram}.
\newblock \bibinfo{journal}{Computer Methods and Programs in Biomedicine} \emph{\bibinfo{volume}{221}}, \bibinfo{pages}{106858}.
%Type = Inproceedings
\bibitem[{Joo et~al.(2023)Joo, Lee, Kim, Jeong and Kim}]{joo2023twelve}
\bibinfo{author}{Joo, M.Y.}, \bibinfo{author}{Lee, H.}, \bibinfo{author}{Kim, J.}, \bibinfo{author}{Jeong, D.S.}, and \bibinfo{author}{Kim, K.S.} (\bibinfo{year}{2023}). \bibinfo{title}{Twelve-lead ecg reconstruction from single-lead signals using generative adversarial networks}.
\newblock In \bibinfo{booktitle}{International Conference on Medical Image Computing and Computer-Assisted Intervention}. \bibinfo{organization}{Springer} pp. \bibinfo{pages}{184--194}.
%Type = Article
\bibitem[{Mason et~al.(2024)Mason, Pandey, Gadaleta, Topol, Muse and Quer}]{mason2024ai}
\bibinfo{author}{Mason, F.}, \bibinfo{author}{Pandey, A.C.}, \bibinfo{author}{Gadaleta, M.}, \bibinfo{author}{Topol, E.J.}, \bibinfo{author}{Muse, E.D.}, and \bibinfo{author}{Quer, G.} (\bibinfo{year}{2024}). \bibinfo{title}{Ai-enhanced reconstruction of the 12-lead electrocardiogram via 3-leads with accurate clinical assessment}.
\newblock \bibinfo{journal}{NPJ Digital Medicine} \emph{\bibinfo{volume}{7}}. \DOIprefix\doi{10.1038/s41746-024-01193-7}.
%Type = Misc
\bibitem[{Gow et~al.(2023)Gow, Pollard, Nathanson, Johnson, Moody, Fernandes, Greenbaum, Waks, Eslami, Carbonati, Chaudhari, Herbst, Moukheiber, Berkowitz, Mark and Horng}]{mimiciv_ecg}
\bibinfo{author}{Gow, B.}, \bibinfo{author}{Pollard, T.}, \bibinfo{author}{Nathanson, L.A.}, \bibinfo{author}{Johnson, A.}, \bibinfo{author}{Moody, B.}, \bibinfo{author}{Fernandes, C.}, \bibinfo{author}{Greenbaum, N.}, \bibinfo{author}{Waks, J.W.}, \bibinfo{author}{Eslami, P.}, \bibinfo{author}{Carbonati, T.}, \bibinfo{author}{Chaudhari, A.}, \bibinfo{author}{Herbst, E.}, \bibinfo{author}{Moukheiber, D.}, \bibinfo{author}{Berkowitz, S.}, \bibinfo{author}{Mark, R.}, and \bibinfo{author}{Horng, S.} (\bibinfo{year}{2023}).
\newblock \bibinfo{title}{Mimic-iv-ecg: Diagnostic electrocardiogram matched subset (version 1.0)}. .
\newblock \URLprefix \url{https://physionet.org/content/mimic-iv-ecg/1.0/}.
%Type = Article
\bibitem[{Li et~al.(2025)Li, Aguirre, Moura, Liu, Zhong, Sun, Clifford, Westover and Hong}]{li2024ecgfounder}
\bibinfo{author}{Li, J.}, \bibinfo{author}{Aguirre, A.}, \bibinfo{author}{Moura, J.}, \bibinfo{author}{Liu, C.}, \bibinfo{author}{Zhong, L.}, \bibinfo{author}{Sun, C.}, \bibinfo{author}{Clifford, G.}, \bibinfo{author}{Westover, B.}, and \bibinfo{author}{Hong, S.} (\bibinfo{year}{2025}). \bibinfo{title}{An electrocardiogram foundation model built on over 10 million recordings with external evaluation across multiple domains}.
\newblock \bibinfo{journal}{NEJM AI} \emph{\bibinfo{volume}{2}}. \DOIprefix\doi{10.1056/AIoa2401033}.
%Type = Misc
\bibitem[{Kingma and Welling(2022)}]{kingma2022autoencodingvariationalbayes}
\bibinfo{author}{Kingma, D.P.}, and \bibinfo{author}{Welling, M.} (\bibinfo{year}{2022}).
\newblock \bibinfo{title}{Auto-encoding variational bayes}. .
\newblock \URLprefix \url{https://arxiv.org/abs/1312.6114}. \href{http://arxiv.org/abs/1312.6114}{\tt arXiv:1312.6114}.
%Type = Article
\bibitem[{Liu et~al.(2020)Liu, Zhang, Li, Wang, Li and Zhang}]{liu2020classification}
\bibinfo{author}{Liu, Y.}, \bibinfo{author}{Zhang, H.}, \bibinfo{author}{Li, J.}, \bibinfo{author}{Wang, Y.}, \bibinfo{author}{Li, Z.}, and \bibinfo{author}{Zhang, X.} (\bibinfo{year}{2020}). \bibinfo{title}{Classification of myocardial infarction with multi-lead ecg signals using deep learning}.
\newblock \bibinfo{journal}{Journal of Electrocardiology} \emph{\bibinfo{volume}{59}}, \bibinfo{pages}{1--7}. \DOIprefix\doi{10.1016/j.jelectrocard.2020.02.004}.
%Type = Article
\bibitem[{Li et~al.(2021{\natexlab{b}})Li, Liu, Zhang, Wang, Li and Zhang}]{li2021classification}
\bibinfo{author}{Li, X.}, \bibinfo{author}{Liu, Y.}, \bibinfo{author}{Zhang, H.}, \bibinfo{author}{Wang, Y.}, \bibinfo{author}{Li, Z.}, and \bibinfo{author}{Zhang, X.} (\bibinfo{year}{2021}{\natexlab{b}}). \bibinfo{title}{Classification of conduction blocks using deep learning on multi-lead ecg signals}.
\newblock \bibinfo{journal}{Journal of Electrocardiology} \emph{\bibinfo{volume}{60}}, \bibinfo{pages}{1--7}. \DOIprefix\doi{10.1016/j.jelectrocard.2021.03.001}.
%Type = Article
\bibitem[{Shukla et~al.(2023)Shukla, Bhatia and Janghel}]{shukla2023arrhythmia}
\bibinfo{author}{Shukla, A.}, \bibinfo{author}{Bhatia, S.}, and \bibinfo{author}{Janghel, R.R.} (\bibinfo{year}{2023}). \bibinfo{title}{Detection of arrhythmia heartbeats from ecg signal using wavelet transform-based cnn model}.
\newblock \bibinfo{journal}{Springer Proceedings in Complexity} \emph{\bibinfo{volume}{2023}}, \bibinfo{pages}{1--12}. \DOIprefix\doi{10.1007/s44196-023-00256-z}.

\end{thebibliography}

\bigskip

%%%  In your References, please include only articles 
%%%  that are published (online publication and 
%%%  preprint servers are OK). Unpublished data, 
%%%  submitted and/or accepted manuscripts, abstracts, 
%%%  and personal communications should be cited within 
%%%  the text only ("unpublished data," "data not 
%%%  shown," "Alice Smith, personal communication") 
%%%  and not included in the references list. Personal 
%%%  communication should be documented by a letter 
%%%  of permission. Whenever possible, please make 
%%%  sure your .bib file has the complete author lists 
%%%  for each item (at minimum, the first 11 authors 
%%%  listed). 

\newpage
\begin{appendices}
\section*{Appendix A: Additional Case Study Samples}
In this appendix, we provide additional case study samples to further illustrate the effectiveness of model-generated ECGs in assisting physicians with myocardial infarction (MI) localization. These samples complement the representative cases presented in the main text and highlight the model's potential to support clinical decision-making by accurately reconstructing infarct regions.

\begin{table}[H]
\centering
\small
\caption{Additional case study samples for model-assisted MI localization.}
\begin{tabular}{cccc}  % 没有竖线
\toprule
\textbf{Physician 1} & \textbf{Physician 2} & \textbf{Physician 3} & \textbf{Model Prediction} \\
\midrule
\makecell{Anterolateral \\ Inferior} & Anterolateral & Anterior & Anterolateral \\
\midrule
Anterolateral & Anteroseptal & No MI & Anteroseptal \\
\midrule
Anteroseptal & Septal & \makecell{Anteroseptal \\ Inferior} & Anteroseptal \\
\midrule
Anterolateral & Anteroseptal & \makecell{Anterior \\ Anteroseptal} & Anteroseptal \\
\midrule
\makecell{Anterolateral \\ Inferior} & Anteroseptal & No MI & Anteroseptal \\
\midrule
Anterolateral & Inferior & Inferior & Inferior \\
\midrule
\makecell{Inferior \\ Lateral} & Inferior & Inferior & Inferior \\
\midrule
Lateral & Lateral & Lateral & \makecell{Inferior \\ Lateral} \\
\midrule
Septal & Septal & Inferior & Septal \\
\bottomrule
\end{tabular}
\label{tab:case_study_additional}
\end{table}

\section*{Appendix B: Additional Reconstruction Examples}

\begin{figure}[H]
    \centering
    \includegraphics[width=0.9\linewidth]{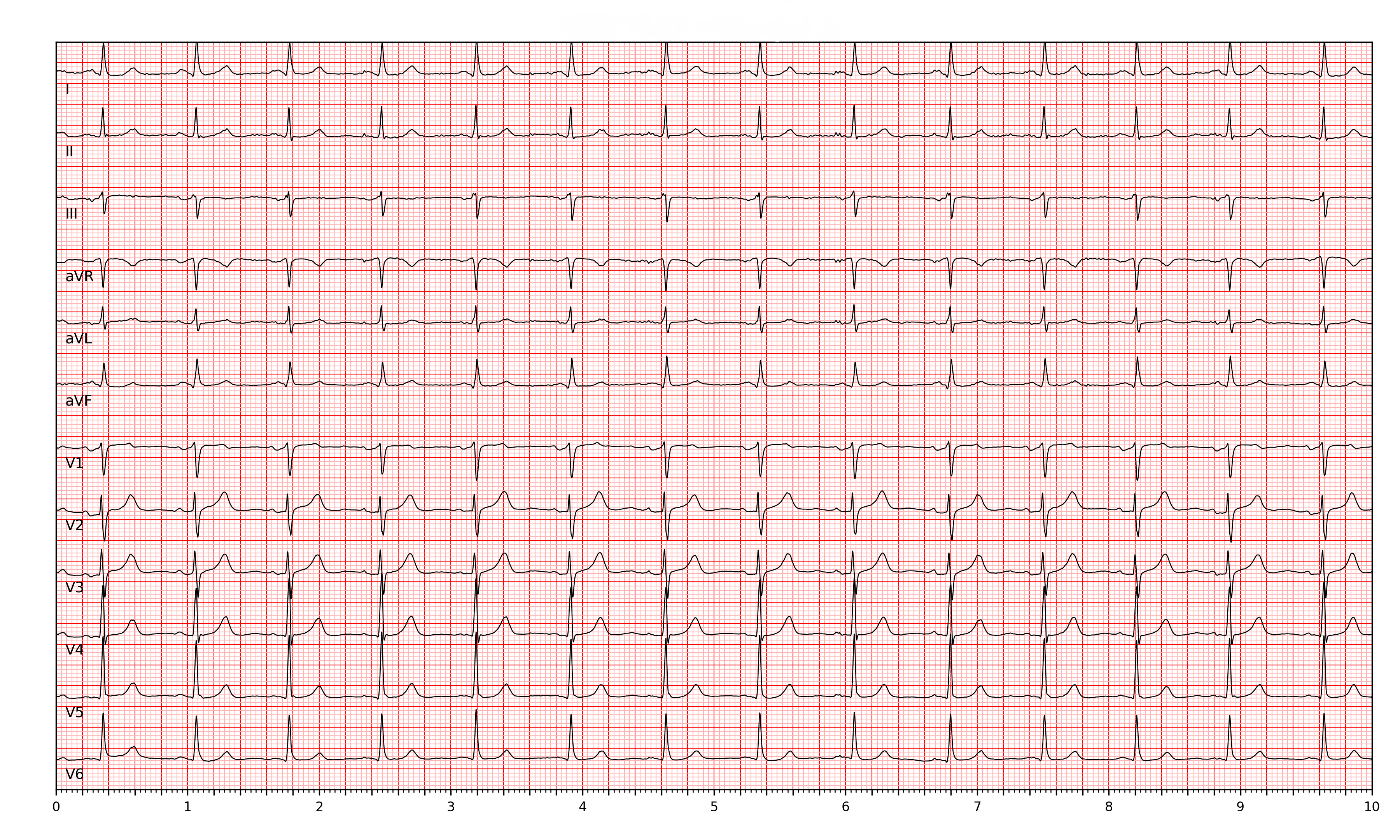}
    \caption{Representative examples of reconstructed 12-lead ECGs from the II/V1/V5 input configuration. Reconstructions closely match the ground truth morphology, including P-wave, QRS complex, and T-wave patterns.}
    \label{fig:recon_examples_appendix}
\end{figure}

Figure~\ref{fig:recon_examples_appendix} shows examples of reconstructed 12-lead ECGs using the II/V1/V5 configuration. Visual inspection indicates that the reconstructions closely resemble the ground truth across different waveform patterns. These examples illustrate the reconstruction quality of the generative framework under a clinically relevant lead setup.

% \section{Prompts for Morphological Feature Extraction of Flow-Volume Curves}\label{appendixA}

% \newpage
% \section{Report Generation Prompt}\label{appendixB}

% \newpage
% \section{Evaluation Prompt for Diagnostic Reports}\label{appendixD}

% \newpage
% \section{Required Fields for Label Extraction.}\label{appendixE}

\end{appendices}

\end{document}